\documentclass{article}
\usepackage{ijcai23}

\usepackage{times}
\usepackage{soul}
\usepackage{url}
\usepackage[hidelinks]{hyperref}
\usepackage[utf8]{inputenc}
\usepackage[small]{caption}
\usepackage{graphicx}
\usepackage{amsmath}
\usepackage{amsthm}
\usepackage{booktabs}
\usepackage{algorithm}
\usepackage{algorithmic}
\usepackage[switch]{lineno}
\usepackage{lipsum}
\usepackage{here}
\usepackage{amsmath,amsthm,amssymb}
\usepackage{graphics}
\usepackage{graphicx}
\usepackage{float}
\usepackage{type1cm} 
\usepackage{tabularx}
\usepackage{booktabs}
\usepackage{subcaption}
\usepackage{thm-restate}
\usepackage{bbm, bm}
\usepackage{mathtools}
\usepackage{cleveref}
\usepackage{caption}
\usepackage{multirow}
\usepackage{fancybox,ascmac}
\usepackage{subcaption}
\newcommand{\indep}{\perp \!\!\! \perp}

\newcommand{\argmin}{\mathop{\rm arg~min}\limits}

\newtheorem{theorem}{Theorem}

\title{Statistically Significant Concept-based Explanation \\of Image Classifiers via Model Knockoffs}

\author{
Kaiwen Xu$^{1,3}$
\and
Kazuto Fukuchi$^{1,3}$\and
Youhei Akimoto$^{1,3}$\And
Jun Sakuma$^{2,3}$
\affiliations
$^1$University of Tsukuba\\
$^2$Tokyo Institute of Technology\\
$^3$RIKEN AIP
\emails
kaiwen@mdl.cs.tsukuba.ac.jp,
\{fukuchi, akimoto\}@cs.tsukuba.ac.jp,
sakuma@c.titech.ac.jp
}

\begin{document}

\maketitle

\begin{abstract}
A concept-based classifier can explain the decision process of a deep learning model by human-understandable concepts in image classification problems. However, sometimes concept-based explanations may cause false positives, which misregards unrelated concepts as important for the prediction task. Our goal is to find the statistically significant concept for classification to prevent misinterpretation. In this study, we propose a method using a deep learning model to learn the image concept and then using the Knockoff samples to select the important concepts for prediction by controlling the False Discovery Rate (FDR) under a certain value. We evaluate the proposed method in our synthetic and real data experiments. Also, it shows that our method can control the FDR properly while selecting highly interpretable concepts to improve the trustworthiness of the model.
\end{abstract}

\section{Introduction}
In recent years, the development of Deep Neural Networks (DNNs) has achieved highly accurate predictions in various tasks. On the other hand, since the prediction process of DNNs is hard to understand for humans, this has driven the development of explainable artificial intelligence (XAI). 

Some post-hoc explanation methods have been proposed to explain the DNNs in the image classification task. For example, methods based on saliency map highlight the pixels relevant to a specific class as visual explanation  \cite{DBLP:conf/cvpr/ZhouKLOT16}. Compared to these post-hoc methods, the concept-based explanation can provide a basis for prediction by high-level concepts that are intuitively comprehensible to humans. For example, when a doctor predicts the severity of arthritis for a patient using a deep learning model, instead of giving the diagnosis result directly, the result is based on concepts like “the possibility of arthritis is high because the joint spacing is narrow”, which is more reliable for humans.

The concept-based explanation is highly expressive; however, false positives often appear that could mislead humans' understanding. A false positive explanation occurs when an unimportant concept is mistakenly identified as important for the prediction task. For example, consider a deep learning model that is trained to classify images of animals. The model may be able to identify a horse in an image correctly, but it may also mistakenly identify the image's background as important for the classification. In this case, the background is an unimportant concept mistakenly identified as important, resulting in a false positive explanation.

In this study, we attempt to suppress the false positive explanations by providing explanations based on statistically significant concepts. We guarantee the reliability of the concept-based explanation by controlling the False Discovery Rate (FDR) of the selected concepts. Our motivation is to decrease the false positive explanations in the concept-based explanation model. We aim to select the statistically significant concepts for the prediction task that can improve the model's reliability.

\subsection{Related Work}
\paragraph{Representation Learning of Images.} Learning latent representation (\textit{i.e.,} concepts) of images can be divided into unsupervised and supervised learning. The unsupervised learning method aims to discover the concepts from data automatically. Among them, the mainstream approach treats unsupervised concept learning as a disentanglement representation learning. The Variation Autoencoder (VAE) framework \cite{DBLP:journals/corr/KingmaW13} is often used for representation learning, many studies aim to add regularized that encourage the model to generate disentangled representations \cite{DBLP:conf/iclr/HigginsMPBGBML17,DBLP:conf/icml/KimM18,DBLP:conf/nips/ChenLGD18,DBLP:conf/aaai/TranFAS22,DBLP:conf/nips/OShaughnessyCCR20,DBLP:journals/corr/abs-1907-07165}. There are also some methods proposed to learn concepts given samples labeled with concepts with supervised learning \cite{DBLP:conf/icml/KohNTMPKL20,DBLP:journals/natmi/ChenBR20,DBLP:conf/cvpr/StammerSK21,DBLP:conf/cikm/KazhdanDJLW20,DBLP:conf/cvpr/YangLCSHW21}. At the same time, some research support both unsupervised and supervised learning methods depending on available dataset  
\cite{DBLP:conf/cvpr/DingXXPYWT20,DBLP:conf/cvpr/0001VSB22,DBLP:conf/iclr/LocatelloTBRSB20}.
\paragraph{Reliability of Explanation.} Although there are many studies related to improving the interpretability of models, studies on the reliability of explanations, $\textit{i.e.,}$ whether we can have confidence in the explanations have not been sufficiently studied. \cite{DBLP:conf/chi/ZhangDL22} proposed a method to improve the robustness of post-hoc explanation \cite{DBLP:conf/cvpr/ZhouKLOT16,DBLP:conf/iccv/SelvarajuCDVPB17}. \cite{DBLP:conf/iclr/BahadoriH21} aims to improve the reliability of the concept-base explanation by removing confounding information of concept based on CBM \cite{DBLP:conf/icml/KohNTMPKL20} and TCAV \cite{DBLP:conf/icml/KimWGCWVS18}. 

Our goal is to improve the reliability of concept-based explanations by providing important concepts for prediction tasks with statistical guarantees. To the best of our knowledge, our work is the first to provide a statistically significant concept-based explanation. 

\subsection{Contribution}
Our research makes the following contributions:
\begin{itemize}
    \item We propose a method that can control the false discovery rate of concepts under a certain value to suppress the false positive explanation.
    \item  To cope with the diverse conceptual learning methods, our method works on both supervised and unsupervised methods for concept learning.
    \item We evaluate our method on both synthetic and real datasets, and the results show that our method can provide highly interpretable concepts while properly controlling the false discovery rate. 
\end{itemize}

\section{Problem Setup}
\subsection{Concept-based Classification}
Concept-based classification is a task to construct a potentially black-box classifier and explain the constructed classifier's decision process through human-interpretable concepts. Let $\mathcal{X} \subset \mathbb{R}^d, \mathcal{C} \subset \mathbb{R}^p, \mathcal{Y} = \{1,2,\dots, K\}$ be the input space of the image, the space of concepts, and the class labels space respectively. Instead of training a classifier $\Omega :\mathcal{X} \to \mathcal{Y}$ directly, a concept-based classifier firstly maps the input space to human-interpretable concepts\footnote{For example, in gender classification from face images, hair length and skin color can be human interpretable concepts.} by $\phi:\mathcal{X} \to \mathcal{C}$ and then use $f:\mathcal{C} \to \mathcal{Y}$ to predict the class labels from the obtained concepts. $\phi$ is a feature extractor built with a neural network and $f$ is a linear model\footnote {$f$ can also be other interpretable prediction models such as decision trees. However, we restrict it to the linear model in our study.}. When the input of $f$ is an explainable feature and since $f$ is a linear model, we can learn which concept is used to predict the class labels by observing the coefficients of $f$. In a linear model, the larger the absolute value of the model coefficient, the more important the corresponding feature is to the prediction. 

We introduce two settings of the concept-based classification, unsupervised and supervised concept-based explanation. Since the unsupervised setting is more elemental, we define the problem of unsupervised concept-based explanation first.
\subsubsection{Unsupervised Concept-based Explanation}
The learner in the unsupervised concept-based explanation has access to the dataset of the input image and class label pairs, $ \{\bm{\mathrm{x}}^{(i)},y^{(i)}\}_{i=1} ^n \in (\mathcal{X}\times\mathcal{Y})^n $. We train our feature extractor $\phi$ to provide meaningful concepts useful for concept-based explanation from the given dataset. 

The Variation Autoencoder (VAE) framework \cite{DBLP:journals/corr/KingmaW13} is often used for concept learning in the unsupervised setting. It consists of two neural networks, an encoder network $\phi$ and a decoder network $\theta$, where the encoder part of VAE $\phi : \bm{\mathrm{x}} \mapsto \bm{\mathrm{z}}$ generates disentanglement representation $\bm{\mathrm{z}} = \phi(\bm{\mathrm{x}})$ of each image as the concepts, and the decoder part $\theta:\bm{\mathrm{z}} \mapsto \hat{\bm{\mathrm{x}}}$ reconstructs image $\hat{\bm{\mathrm{x}}}$ as close as possible to the input image $\bm{\mathrm{x}}$. We can obtain the entire network by optimizing the following function
\begin{equation}
    \mathcal{L}_{\text{VAE}} = \mathcal{L}_{R}(\bm{\mathrm{x}},\hat{\bm{\mathrm{x}}}) + \alpha \mathcal{L}_{D}(\bm{\mathrm{z}},\mathcal{N}(\bm{0},\bm{I})).
    \label{L_VAE}
\end{equation}
Here, $\mathcal{L}_R(\cdot,\cdot)$ measures the reconstruction loss between $\bm{\mathrm{x}}$ and $\hat{\bm{\mathrm{x}}}$ and $\mathcal{L}_D(\cdot,\cdot)$ measures the degree of independence between the concepts, which encourages the concepts to capture independent and interpretable factors of image.

\subsubsection{Supervised Concept-based Explanation}
The learner in the supervised concept-based explanation can access the ground truth label of the concept on top of the dataset used in unsupervised learning methods. Specifically,  the dataset is given by  $\mathcal{D} = \{\bm{\mathrm{x}}^{(i)},\bm{\mathrm{c}}^{(i)},y^{(i)}\}_{i=1} ^n$, where $\bm{\mathrm{c}}^{(i)} \in \mathcal{C}$ represents ground truth concepts of each image. For example, when $\bm{\mathrm{x}}$ represents a face image, then $\bm{\mathrm{c}}$ can represent concepts such as hair color or whether there is a beard. 

Concept Bottleneck Model (CBM) \cite{DBLP:conf/icml/KohNTMPKL20} is one kind of supervised concept-based explanation method. The objective function of CBM can be expressed as :
\begin{equation}
    \mathcal{L}_{\text{CBM}} =  \mathcal{L}_{C} (\bm{\mathrm{z}},\bm{\mathrm{c}}) + \alpha \mathcal{L}_{Y}(y,\hat{y}).
    \label{L_CBM}
\end{equation}
Here, $\mathcal{L}_C(\cdot,\cdot)$ represents the supervised loss of concepts which encourages the concepts $\bm{\mathrm{z}} = \phi(\bm{\mathrm{x}})$ predicted with the encoder to be as close as possible to the concept labels $\bm{\mathrm{c}}$. $\mathcal{L}_Y(\cdot,\cdot)$ measures the prediction loss between predicted target $\hat{y} = f(\bm{\mathrm{z}})$ and ground truth class label.

\subsection{Concept Selection with Controlling FDR}
To obtain a concept-based explanation, we investigate the coefficients of linear model $f$ and select the concepts with large coefficients as important concepts for prediction. However, the coefficients of the linear model are affected not only by the association with the class labels but some other various factors, such as noise or correlation between input variables.  Therefore, the concepts selected by only investigating the coefficients of a linear model may not be really important for prediction, which leads to low reliability. To improve reliability, we need to control the false discovery rate of unimportant concepts below a certain value.

Let   $\mathcal{H}_0 \subseteq \{1,\dots,p\}$ be the set that contains all indices corresponding to all unimportant variables (\textit{i.e.,} concepts) in $\bm{\mathrm{z}}$, where we will give its specific definition later in Section 3. Our goal is to find the largest subset $\hat{\mathcal{S}} \subseteq \{1,\dots,p\}$ that contains important concepts while controlling the FDR under a certain value $q$. The FDR is defined as:
\begin{equation}
    \label{FDR}
    \text{FDR} = \mathbb{E}\left [ \frac{\arrowvert \hat{\mathcal{S}} \cap \mathcal{H}_0\arrowvert}{\max (1,\arrowvert \hat{\mathcal{S}}\arrowvert)}\right].
\end{equation}
Here, FDR represents the proportion of unimportant concepts identified as important mistakenly. If we guarantee the upper bound of FDR of the concept selection results, we can say our explanation is reliable in the sense of statistical significance.
\section{Concept Selection via Knockoff}
To address the problem mentioned in Section 2.2, we propose to employ the technique of the Knockoff filter \cite{candes2018panning}. Knockoff is a variable selection method that controls the FDR under the linear regression setup. In this section, we first describe the Knockoff filter for variable selection and then propose a methodology to combine the Knockoff filter with a concept-based explanation.
\subsection{Knockoff}
Let $X = (X_1,\dots,X_p) \in \mathbb{R}^p$ be the explanatory variable and $Y \in \mathbb{R}$ be the response variable.   Here, we assume there exists a linear model $:Y = \bm{\beta} ^\top X + \epsilon$ where $\bm{\beta}$ is a sparse vector whose non-zero elements represent the ground truth of features important to $Y$, and $\epsilon$ represents the noise. Define the unimportant variables set as $\mathcal{H}_0 = \{ j:\beta _j = 0\}$
. Then, the goal of the Knockoff filter is to select the largest subset of variables $\hat{\mathcal{S}}$ such that FDR in Eq. (\ref{FDR}) is controlled under a certain value $q$.

\subsubsection{Model-X Knockoff}
\cite{candes2018panning} proposed to generate a Knockoff sample $\Tilde{X} \in \mathbb{R}^p$ of $X$, which satisfies the following properties
\begin{align}
\label{modelx_d1}
(X,\tilde{X})_{\text{swap}(S)} &\overset{d}{=}(X,\tilde{X}), \\
\label{modelx_d2}
\tilde{X} &\indep  Y \arrowvert X.
\end{align}

Here, $S$ can be any subset of $\{1,\dots,p\}$ and the symbol $\overset{d}{=}$ means equality in distribution.  Eq. (\ref{modelx_d1}) requires that the joint distribution will be invariant when any subset of variables is swapped with their Knockoff. For example when $p=3$ and $S=\{2,3\}$, we require that  $(X_1,X_2,X_3,\tilde{X}_1,\tilde{X}_2,\tilde{X}_3)$ has the same distribution as $(X_1,\tilde{X}_2,\tilde{X}_3,\tilde{X}_1,X_2,X_3)$. Eq. (\ref{modelx_d2}) requires that the Knockoff sample $\tilde{X}$ is generated without looking at $Y$.

Once Knockoff has been generated (the way to generate Knockoff samples 
will be explained later), we calculate the statistics $W_j$ for each 
variable $X_j$ where $j \in \{1,\dots,p\}$. Here, for $n$ observations, we make $\bm{Y} \in \mathbb{R}^n$ to represent the observation of response variable and  $[\bm{X},\tilde{\bm{X}}] \in \mathbb{R}^{n\times 2p}$ represents an augmented matrix of the original variable with their Knockoffs. The common choice to calculate the $W_j$ is using L1-regularization as the following:
\begin{align}
    \label{modelx_1}
    \hat{\bm{b}} &= \argmin _{\bm{b} \in \mathbb{R}^{2p}} \frac{1}{2}\Arrowvert\bm{Y}-[\bm{X},\tilde{\bm{X}}]\bm{b}\Arrowvert_2 ^2 + \lambda \Arrowvert\bm{b}\Arrowvert _1 ~ ,\\
    \label{modelx_2}
    W_j &= \arrowvert \hat{b}_j\arrowvert - \arrowvert \hat{b}_{j+p} \arrowvert .
\end{align}
Here $b_{1:p}$ and $b_{p+1:2p}$ correspond to the parameters of the original variable and  Knockoffs, respectively. Recall that Knockoffs are synthetic variables constructed to be correlated with the original variables but are not predictive of the target variable. By comparing the magnitude of the coefficients of the model when trained on the original features with the Knockoff, we can determine which variables are important for predicting the target variable. A large and positive $W_j$ can be the evidence against the hypothesis that $X_j$ is unimportant for $Y$. Then the FDR can be controlled under a certain value $q$ by the following theory.
\begin{theorem}\label{thm:model-x-fdr}\cite{candes2018panning}
Given $\bm{Y}$ and $\bm{X}$ following the linear model, let $\tilde{\bm{X}}$ be the Knockoff sample satisfying Eq. (\ref{modelx_d1}) and Eq. (\ref{modelx_d2}). Suppose $W_j$ is calculated by Eq. (\ref{modelx_2}) from $\bm{Y}, \bm{X}$, and $\tilde{\bm{X}}$. Given a target FDR level $q$, we select variables using $W_j$ as $\hat{\mathcal{S}} = \{j:W_j \geq \tau\}$, where 
\begin{equation*}
    \tau = \min \left \{ t>0 : \frac{1 + \arrowvert\{j:W_j \leq -t\}\arrowvert}{\arrowvert\{j:W_j \geq t\}\arrowvert} \leq q \right \}.
\end{equation*}
Then, the FDR of $\hat{\mathcal{S}}$ is controlled as the prescribed level, i.e.,
 \begin{equation*}
    \mathbb{E}\left [\frac{\arrowvert \hat{\mathcal{S}} \cap \mathcal{H}_0\arrowvert}{\max (1,\arrowvert \hat{\mathcal{S}}\arrowvert)}\right ] \leq q .
\end{equation*}
\label{theorem1}
\end{theorem}
\paragraph{Second-order Sampler.}
Generating Knockoff sample that strictly achieves Eq (\ref{modelx_d1}) is difficult in practice. \cite{candes2018panning} introduced a relaxation called the second-order sampler to generate the Knockoff sample. Unlike Eq. (\ref{modelx_d1}), this method only needs to match the first two moments of distributions. So when $X$ is a multivariate Gaussian, second-order Knockoff \cite{candes2018panning} works properly since matching the first two moments can exactly satisfy the Eq. (\ref{modelx_d1}). 

\paragraph{Deep Knockoff Sampler.}
When $X$ does not follow as a multivariate Gaussian, Theorem \ref{theorem1} does not hold only by matching the first two moments. \cite{romano2020deep} proposed to introduce a generative model $f_{\text{deep}}:\mathbb{R}^p \to \mathbb{R}^p$ to generate the  Knockoff sample $\tilde{X}$ which approximately satisfies Eq. (\ref{modelx_d1}). 

\subsection{Concept Selection via Model Knockoff}
In this subsection, we introduce how to apply the Knockoff filter to the concept selection of image classification tasks. 

The concept selection process is summarized in Algorithm \ref{A1}. We divided the dataset into two parts, the dataset $\mathcal{D}_{L}$ for feature extractor training and the dataset $\mathcal{D}_{S}$ for concept selection as shown in step 1. First, we train a feature extractor on $\mathcal{D}_L$ and obtain the learned feature extractor $\hat{\phi}$. We remark that the training procedure of the feature extractor is chosen depending on the dataset the learner has access to, as we already discussed in Section 2.1. Then, we apply $\hat{\phi}$  to $D_S$ to obtain concepts $\bm{\mathrm{z}}^{(i)}=\hat{\phi}(\bm{\mathrm{x}}^{(i)})$  for each image on $\mathcal{D}_S$ (step 3 - step 7). After that, we generate the Knockoff sample $\Tilde{\bm{\mathrm{z}}}$ of each concept $\bm{\mathrm{z}}$ and obtain the dataset $\{\bm{\mathrm{z}}^{(i)},\Tilde{\bm{\mathrm{z}}}^{(i)},y^{(i)}\}_{i=1} ^{\arrowvert \mathcal{D}_S \arrowvert}$ (step 8). Then, we construct the linear model of $[\bm{\mathrm{z}},\tilde{\bm{\mathrm{z}}}] \to y$ using the method described in Section 3.1 and compute the statistics to obtain the concept selection result  $\hat{\mathcal{S}}$ with a given FDR level (step 9 - step 13).


\begin{algorithm}[tb]
    \caption{Algorithm for concept selection}
    \label{alg:algorithm}
    \textbf{Input}: Dataset: $\mathcal{D}$, FDR: $q$\\
    \textbf{Output}: selected concept set: $\hat{\mathcal{S}}$
    \begin{algorithmic}[1] 
        \STATE Split the data into two parts, $\mathcal{D}_{L}$ for concept learning and $\mathcal{D}_{S}$ for concept selection.
        \STATE Optimize the $\hat{\phi}$ on $\mathcal{D}_{L}$. 
        \STATE $\bm{\mathrm{Z}} \gets \{\}$
          \FOR {$i$ in $\{1,2,\dots,\arrowvert \mathcal{D}_S \arrowvert\}$}
          \STATE $\bm{\mathrm{z}}^{(i)} \gets \hat{\phi}(\bm{\mathrm{x}}^{(i)})$ //generate concepts for each image.
          \STATE $\bm{\mathrm{Z}} \gets \bm{\mathrm{Z}} \cup \bm{\mathrm{z}}^{(i)} $
          \ENDFOR
        \STATE Generate the Knockoff sample $\widetilde{\bm{\mathrm{Z}}} = \{\tilde{\bm{\mathrm{z}}}^{(1)},\dots,\tilde{\bm{\mathrm{z}}}^{(\arrowvert \mathcal{D}_S \arrowvert)}\}$ of $\bm{\mathrm{Z}}$.
        \STATE Obtain the $\hat{\bm{b}}$ as Eq. (\ref{modelx_1}).
        \STATE Calculate the statistic $W_j$ as Eq. (\ref{modelx_2}).
        \STATE Calculate the threshold $\tau$ by Theorem (1).
        \STATE $\hat{\mathcal{S}} \gets  \{j:W_j \geq \tau\}.$
        \STATE \textbf{return} $\hat{\mathcal{S}}$
    \end{algorithmic}
    \label{A1}
\end{algorithm}

\subsection{FDR Control in Knockoff-based Concept Selection}
This subsection demonstrates that Algorithm \ref{A1} adequately controls FDR below the prescribed level. Algorithm \ref{A1} carries out the Knockoff filter using the concept $\hat\phi(x)$ and class label $y$ as the explanatory and response variables, respectively. This process implicitly supposes the existence of the underlying linear model between $\hat\phi(x)$ and $y$. The important concepts that should be selected are the non-zero elements of the underlying model's parameter. Such an underlying linear model is uniquely determined to correspond to the feature extractor $\hat\phi$ under a mild condition. The next theorem shows the closed form of the underlying linear model and the FDR control accomplished by Algorithm \ref{A1}.
\begin{theorem}\label{thm:beta-phi}
    Let $\hat{\mathcal{S}}$ be the selected concepts by Algorithm \ref{A1} with an arbitrary concept-based explanation method to construct $\hat\phi$. Suppose that Algorithm \ref{A1} employs Model-X Knockoff with the FDR level $q \in (0,1)$ to generate the Knockoff sample. Also, suppose that $\mathbb{E}[\hat\phi(X)\hat\phi(X)^\top]$ is positive definite. Then, $\hat{\mathcal{S}}$ satisfies $\mathrm{FDR} \le q$ for $\mathcal{H}_0=\{j : \beta_{\phi,j} = 0\}$ where
    \begin{align}
        \beta_\phi = \mathbb{E}\left[\hat\phi(X)\hat\phi(X)^\top\right]^{-1}\mathbb{E}\left[\hat\phi(X)Y\right]. \label{eq:beta-phi}
    \end{align}
\end{theorem}


 

The proof of Theorem \ref{thm:beta-phi} is left to the supplementary material. As proved by Theorem \ref{thm:beta-phi}, Algorithm \ref{A1} adequately controls the FDR below $q$ for $\mathcal{H}_0$ defined in Theorem \ref{thm:beta-phi}. In Eq. (\ref{eq:beta-phi}), $\mathbb{E}[\hat\phi(X)Y]$ stands for the direct effect from a concept to the response variable, and $\mathbb{E}[\hat\phi(X)\hat\phi(X)^\top]^{-1}$ stands for the correlation between concepts. Hence, each element of $\beta_\phi$ represents the combination of the direct and indirect effects from a concept to the response variable. An unimportant concept $j$ such that $\beta_{\phi,j} = 0$, is a concept that has neither direct nor indirect effects on the response variable.

\section{Concept Sparsity Regularization}
\paragraph{Difficulty.} As shown in Section 3, the Model-X Knockoff regards a concept as unimportant only if $\beta_{\phi,j} = 0$. However, such a condition may be easily unsatisfied even for unimportant concepts since we train the feature extractor with data, and all concepts depend more or less on the response variable. As the experiments result shown in Fig. \ref{exp2}, which we will discuss later, the concept selection via the Model-X Knockoff combined with the standard concept learning algorithm, such as VAE and CBM,  regards above $80\%$ of all concepts as important. This demonstrates that even with Knockoff, a large number of unimportant concepts for prediction are still selected as important. This leads to decreasing in interpretability and trustworthiness of the explanation by our model.

\paragraph{Solution.} From the interpretability perspective, we want to describe the response variable with a relatively small number of concepts. To this end, we introduce a concept learning algorithm so that the linear model trained with the resulting concepts (features) becomes sparse, that is, $\bm{\beta}_\phi$ is sparse. To make $\bm{\beta} _\phi$ sparse, we construct an estimator of $\bm{\beta} _\phi$ as $\hat{\bm{\beta}}_\phi = \mathop{\rm arg~min}_{\bm{\beta} _\phi} \mathcal{L}_Y (\hat{y},y),$ where $\hat{y} = \bm{\beta}_{\phi}^\top \bm{\mathrm{z}}$ is the prediction of the linear model. Because we expect that  $\hat{\bm{\beta}}_\phi $ well approximates $\bm{\beta} _\phi$ by definition, if we could make $\hat{\bm{\beta}}_\phi$ sparse, $\bm{\beta}_\phi$ would also become sparse as we desired. To achieve this,  we propose to add a regularization term called Concept Sparsity Regularization (CSR) to encourage feature sparsity in the linear model. The CSR is attained by adding the L1 norm of $\hat{\bm{\beta}}_\phi$ to the objective function of the feature extractor learning part, which penalizes $\hat{\bm{\beta}} _\phi$ for having too many non-zero coefficients. Our objective function is defined as:
\begin{align}
    \hat{\phi} &=\argmin _{\phi}  \alpha_1 \mathcal{L}_{R}(\bm{\mathrm{x}},\hat{\bm{\mathrm{x}}}) + \alpha _2 \mathcal{L}_{D}(\bm{\mathrm{z}},\mathcal{N}(\bm{0},\bm{I})) \label{L_all}\\&+ \alpha _3 \mathcal{L}_C (\bm{\mathrm{z}},\bm{c}) + \alpha _4 \arrowvert \hat{\bm{\beta}}_{\phi} \arrowvert _1 , ~s.t.~ \hat{\bm{\beta}}_\phi = \argmin _{\bm{\beta} _\phi} \mathcal{L}_Y (\hat{y},y). \nonumber
\end{align}
\begin{figure*}[t]
\begin{center}
\includegraphics[width=.9\textwidth]{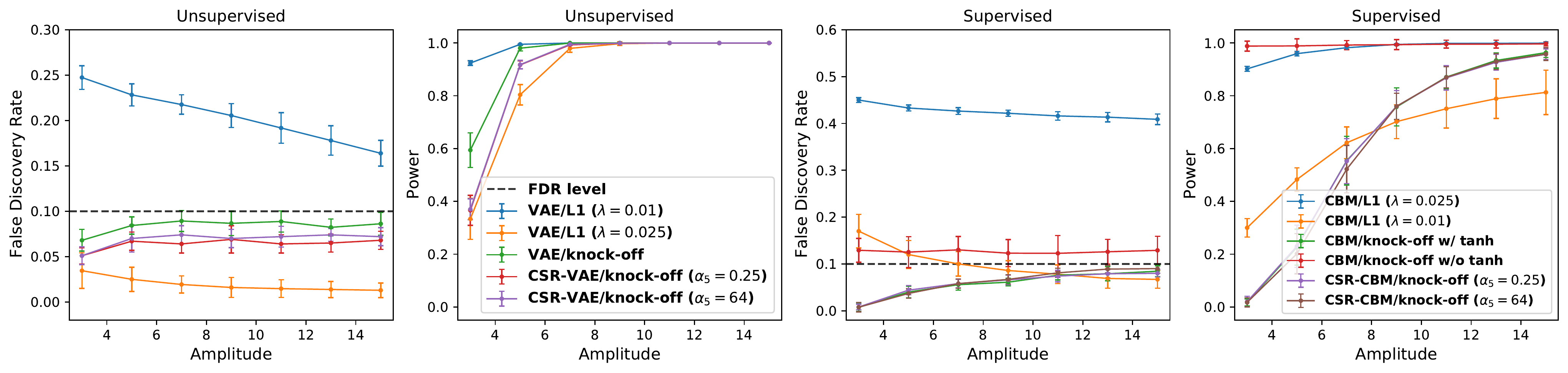}
\end{center}
\caption{Experiments results on synthetic data in unsupervised and supervised settings. The results are averaged under 10 independent trials. }
\label{exp1}
\end{figure*}
\paragraph{Implementation.} Eq. (\ref{L_all}) shows the objective function we want to optimize. However, the formulation requires solving a bi-level optimization problem. This problem requires to optimize $\hat{\bm{\beta}}_{\phi}$ on a linear model while $\hat{\bm{\beta}}_\phi$ depends on feature extractor $\phi$. At the same time, $\phi$ needs to be optimized given $\hat{\bm{\beta}}$,  which can be challenging and time-consuming. As a surrogate of the problem defined by Eq. (\ref{L_all}), we train the parameter $\phi$ and $\bm{\beta}_{\phi}$ jointly using the following objective function:
\begin{align}
    \hat{\phi},\hat{\bm{\beta}}_\phi &= \argmin _{\phi,\bm{\beta}_{\phi}}  \alpha_1 \mathcal{L}_{R}(\bm{\mathrm{x}},\hat{\bm{\mathrm{x}}}) + \alpha _2 \mathcal{L}_{D}(\bm{\mathrm{z}},\mathcal{N}(\bm{0},\bm{I})) \nonumber \\&+ \alpha _3 \mathcal{L}_{C} (\bm{\mathrm{z}},\bm{c})+ \alpha _4 \mathcal{L}_{Y}(y,\hat{y}) + \alpha _5 \arrowvert \bm{\beta}_{\phi} \arrowvert _1 .\label{L_ours}
\end{align}
The problem defined by Eq. (\ref{L_ours}) is reduced to the original optimization problem by taking hyperparameter $\alpha _4 \to \infty$.

Formulation of Eq. (\ref{L_ours}) can be regarded as a generalization of the feature (concept) extraction part of various concept-based classification methodologies shown in Table \ref{summary}. The most basic feature extractor by $\beta-$VAE \cite{DBLP:conf/iclr/HigginsMPBGBML17} is obtained by setting $\alpha_1, \alpha_2 > 0, \alpha_3=0$. If the concept's label is available, we add the supervised loss ($\textit{i.e.}, \alpha _3 > 0$), and it is equal to Full-VAE \cite{DBLP:conf/iclr/LocatelloTBRSB20} which is one kind of supervised VAE. Also, letting $\alpha _1, \alpha _2$ to zero and only retaining $\alpha _3 > 0$, we obtain CBM \cite{DBLP:conf/icml/KohNTMPKL20} which learns the concepts only by annotation data. In our unsupervised setting, we do not have labeled concepts in the dataset, so we just set $\alpha_3 = 0$ in Eq. (\ref{L_all}) called CSR-VAE. In our supervised setting, since we can directly learn the concepts by concepts' ground truth, we set $\alpha_1, \alpha_2 = 0, \alpha_3 > 0$ to perform the concept learning, which is called CSR-CBM.

The difference in our concept selection process from  Algorithm \ref{A1} is in step 2. In this step, we use CSR-VAE (in the unsupervised setting) and CSR-CBM (in the supervised setting) instead of VAE and CBM to make $\bm{\beta}_\phi$ sparse. 

Regarding the FDR guarantee, since Theorem \ref{thm:beta-phi} guarantees FDR regardless of the learning method of $\phi$, it guarantees FDR even for those generated by Eq. (\ref{L_ours}), which is expected to make $\bm{\beta}_\phi$ sparser.


\begin{table}
    \centering
    \resizebox{.5\textwidth}{!}{
    \begin{tabular}{l|lrrrrr}
        \toprule
       setting & concept learning method  & $\alpha _1$ & $\alpha _2$ & $\alpha _3$ & $\alpha _4$& $\alpha _5$ \\
        \midrule
       \multirow{2}{*}{unsupervised} & $\beta-$ VAE [Higgins+17]   &   $\surd$  & $\surd$  & & & \\
        & CSR-VAE (ours) & $\surd$& $\surd$& &$\surd$ &$\surd$\\
        \midrule
        \multirow{3}{*}{supervised} & CBM  [Koh+20]          &            &          & $\surd$ &$\surd$ &    \\
        & Full-VAE [Locatello+20]&   $\surd$  & $\surd$  & $\surd$ & &  \\
        & CSR-CBM (ours) &  &  &$\surd$ & $\surd$ &$\surd$\\
        \bottomrule
    \end{tabular}
    }
    \caption{Various concept learning methodologies covered Eq. (\ref{L_ours}), $\surd$ represents that set the $\alpha _j > 0$.}
    \label{summary}
\end{table}

\section{Experiments on Synthetic Data}
We evaluate whether our method can control the FDR strictly. It is hard to determine which concepts are important for target prediction in real image datasets, leading to the precise evaluation of FDR becoming difficult. Therefore, in this section, we design an image classification problem based on an artificially generated sparse linear model to evaluate whether our method can control the FDR strictly. 
\begin{figure*}
    \centering
  \begin{subfigure}{.23\textwidth}
    \centering
    \includegraphics[width=.95\linewidth]{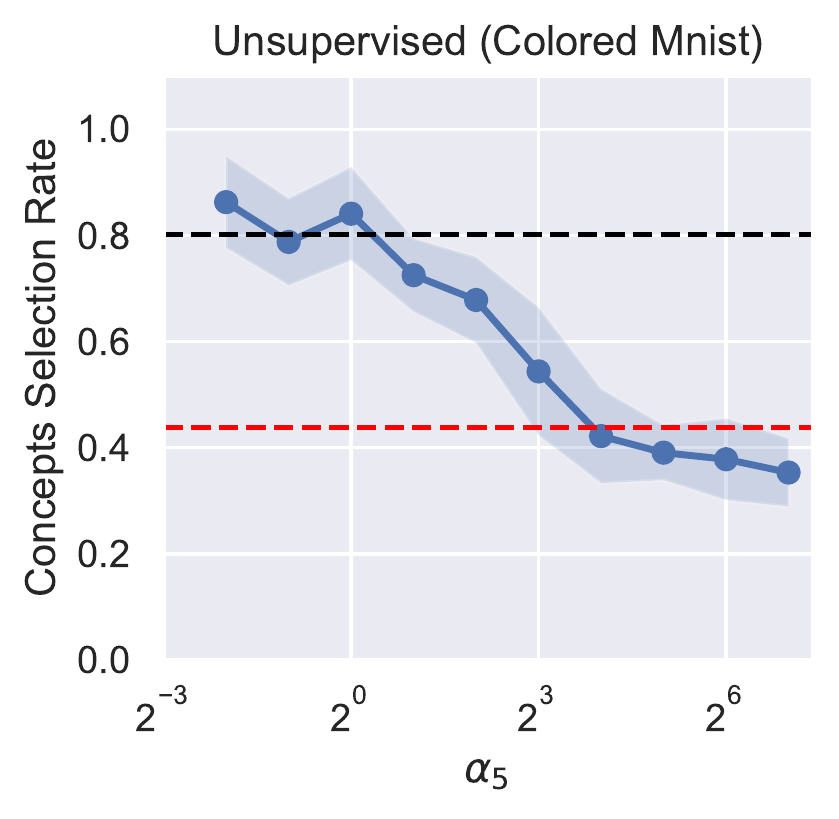}
    \subcaption{Concept selection rate in unsupervised setting}
    \label{exp2_1}
  \end{subfigure}%
  \hfill
  \begin{subfigure}{.23\textwidth}
   \centering
    \includegraphics[width=.95\linewidth]{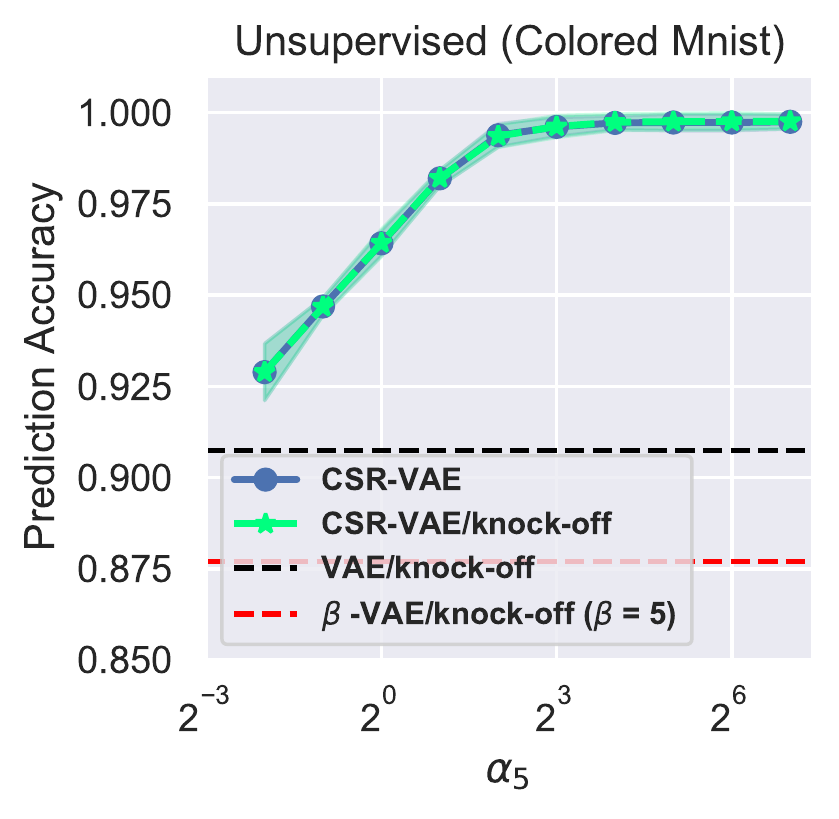}
    \subcaption{Prediction accuracy in unsupervised setting}
    \label{exp2_2}
  \end{subfigure}
  \hfill
  \begin{subfigure}{.23\textwidth}
   \centering
    \includegraphics[width=.95\linewidth]{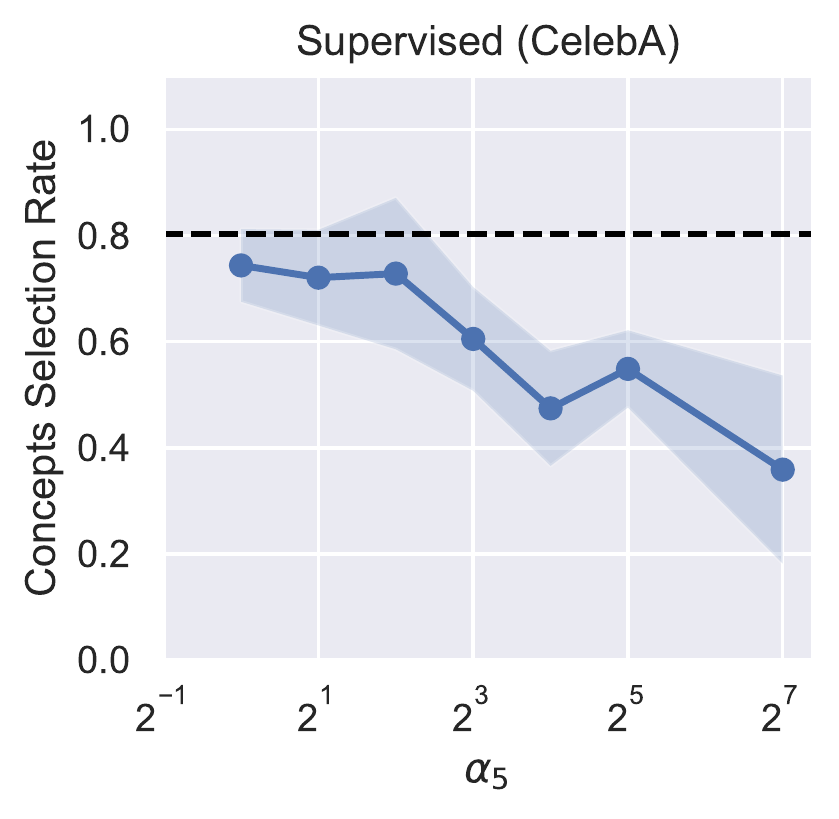}
    \subcaption{Concept selection rate in supervised setting}
    \label{exp2_3}
  \end{subfigure}
  \hfill
  \begin{subfigure}{.23\textwidth}
   \centering
    \includegraphics[width=.95\linewidth]{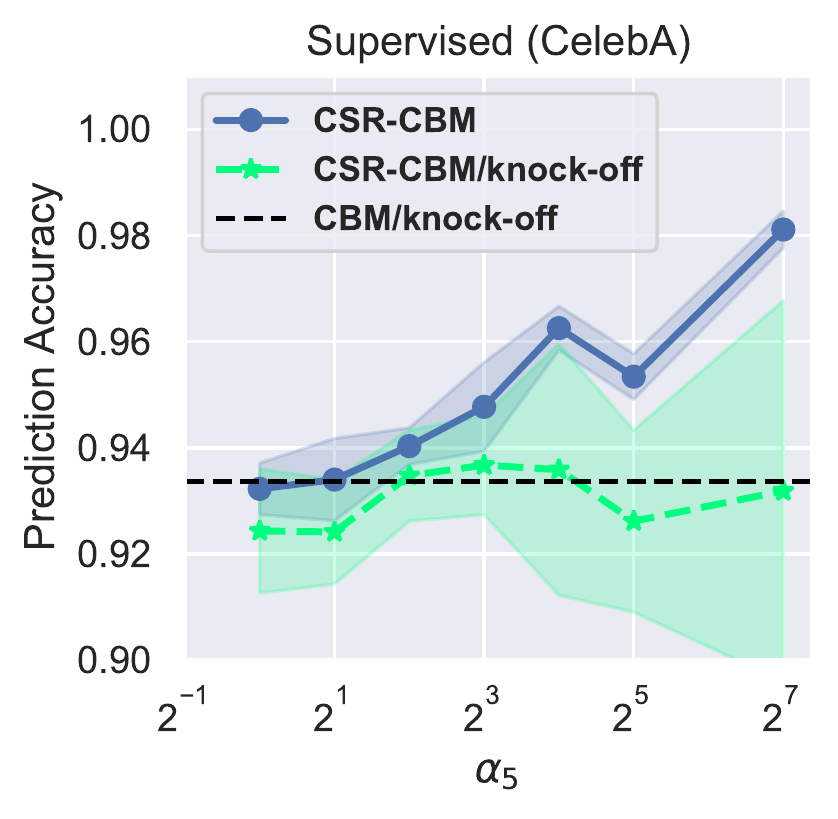}
    \subcaption{Prediction accuracy in supervised setting}
    \label{exp2_4}
  \end{subfigure}
  \caption{Experiments results on real concepts in unsupervised and supervised settings. The results are averaged under 10 independent trials.}
  \label{exp2}
\end{figure*}

\subsection{Experiment Setting}
\paragraph{Dataset.}In our synthetic experiments, we use the CelebA dataset \cite{liu2015faceattributes}, which includes 202,599 face images, to evaluate our method. In the CelebA, each image contains $40$ binary concept annotations (\textit{e.g.,} smile, beard, etc.). In the unsupervised setting, the binary concept annotations are not used. Instead, the concepts are obtained using data in an unsupervised manner. In the supervised learning setting, the binary concept annotations are given, which are used to obtain feature vectors associated with each concept.
\paragraph{Synthetic Data.} We construct a classification problem defined with an artificially defined linear model. Suppose we already obtained a set of concepts $\bm{\mathrm{Z}}$ in the supervised or unsupervised manner. We first randomly sample $m$ concepts $\bm{\mathrm{z}}^{(i)}$, then we make artificial class label $y$ as $y ^{(i)} = \max (0,\text{sgn}(\bm{\beta}^\top \bm{\mathrm{z}}^{(i)} + \epsilon ^{(i)})$, where $\bm{\beta} \in \mathbb{R}^p$ is a sparse vector which contains $k$ randomly chosen non-zero elements equal to $\pm a/\sqrt{m}$ (where the sign is chosen with probability 1/2 and $a\in \mathbb{R}$ represents the signal amplitude); $\epsilon ^{(i)}$ is an $i.i.d$ sample drawn from $ \mathcal{N}(0,1)$.  Here, the concepts associated with non-zero elements in $\bm{\beta}$ are ground truth important concepts for this classification. A small signal amplitude misleads the model to regard the ground truth as noise, and as the signal amplitude increases, the difficulty of ground truth detection will decrease. In this setting, the ground truth of unimportant concepts set is $\mathcal{H}_0 = \{ j:\beta _j = 0\}$ and defined the important concepts set are $\mathcal{H}_1 = \{ j:\beta _j \neq 0\}$.  
\paragraph{Evaluation Metric.}
The evaluation metric is FDR and power. The definition of FDR is stated as Eq. (\ref{FDR}), and the power is defined as
\begin{equation}
     \text{power} = \mathbb{E}\left [\frac{\arrowvert \hat{\mathcal{S}} \cap \mathcal{H}_1\arrowvert}{\arrowvert\mathcal{H}_1\arrowvert}\right ],
     \label{power}
\end{equation}
The expected value in both Eq. (\ref{FDR}) and Eq. (\ref{power}) can be calculated over the randomness of sampling $\{\bm{\mathrm{z}}^{(i)},\tilde{\bm{\mathrm{z}}}^{(i)}\}_{i=1} ^m$.


\paragraph{Concept Selection.} We compare two concept selection procedures. One is L1-regularized logistic regression (termed L1) which regards concepts that give non-zero coefficients as important and do not guarantee FDR. The other is Algorithm \ref{A1} with Knockoff (termed Knockoff), which guarantees FDR as we proved by Theorem \ref{thm:beta-phi}. In the unsupervised concept learning setting, since $\bm{\mathrm{z}}$ generated by the VAE-based method approximately follows the Gaussian distribution, we use the second-order Knockoff sampler to generate the Knockoff samples. While if the concepts are generated by the supervised learning method, since the distribution of $\bm{\mathrm{z}}$ is not properly controlled, we need to train a deep-Knockoff machine priorly to generating Knockoff samples. To facilitate the training of deep-Knockoff, we apply a hyperbolic tangent function to bound $\bm{\mathrm{z}}$ into range $[-1,1]$ to properly bound the distribution of $\bm{\mathrm{z}}$.

\paragraph{Comparison Method.} In the unsupervised learning setting, we compare the combination of two concept learning methods (VAE,  CSR-VAE) and two concept selection methods (L1,  Knockoff). Similarly, in the supervised learning setting, we compare the combination of two concept learning methods (CBM,  CSR-CBM) and two concept selection methods (L1,  Knockoff). We use the notation: concept learning method/concept selection method to distinguish the different methods (\textit{e.g.,} VAE-CSR/Knockoff). In this experiment, we mainly compare the difference between Knockoff and L1 regularization in terms of feature selection. We do not validate the sparsity of CSR because $y$ has been defined as artificially generated with a sparse vector $\bm{\beta}$.

\subsection{Experiment Results}
Fig. \ref{exp1} shows the variable selection results. The horizontal axes represent the value of $a$, which is the signal amplitude, and the vertical axes represent the FDR and power, respectively, with different variable selection methods.  The black dotted line in the FDR graph represents the value of $q$, which we want to control. 

The result shows that concept selection using Knockoff can effectively control the FDR under $q=0.1$ in both unsupervised and supervised settings. The power is also improved by increasing the with signal amplitude $a$. On the other hand, L1-regularization (VAE/L1) cannot control the FDR under $q$ appropriately in most cases, and a lot of false positives happen when $\lambda$ is small. While the reason there is no significant difference when CSR is used or not is because $\bm{\beta}$ is already defined as a sparse vector.


\section{Experiments on Real Concept Data}
In the synthetic data experiments, we show that $\{$VAE, CSR-VAE, CBM, CSR-CBM$\}$/Knockoff can select important concepts by controlling the FDR properly. In this part, we apply our method to image classification of real datasets with and without concept labels and access if $\{$CSR-VAE, CSR-CBM$\}$/Knockoff can select a small number of concepts without knowing the ground truth concepts.
 
\subsection{Dataset}
\paragraph{Colored MNIST.} We manually add six types of colors to the MNIST dataset 
\cite{lecun1998gradient} to create the Colored-MNIST dataset. This dataset is used to evaluate our method in the unsupervised learning setting. The prediction target in this task is digits.

\paragraph{CelebA.} The concept labeled CelebA is also used in our real concept experiments for the supervised learning settings. The concepts other than gender in CelebA are treated as concepts, and gender is treated as the class label that we want to predict with concepts.

\subsection{Sparsity Evaluation}
\paragraph{Settings.} 
We applied $\{$VAE, CSR-VAE, CBM, CSR-CBM$\}$/Knockoff to select concepts important for the image classification task. We control the concept sparsity level of concept selection with CSR by adjusting $\alpha _5$ and compare the number of selected concepts to concept selection without using CSR. We also compare the prediction accuracy of the linear model and the meaning of selected concepts. 

\subsubsection{Results of Concept Sparsity} 
Fig. 2 shows concept selection rate and prediction accuracy of CSR-$\{$VAE,CBM$\}$/Knockoff and $\{$VAE,CBM$\}$/Knockoff. The horizontal axes represent the regularization parameter. The vertical axes represent the concept selection rate (Fig. \ref{exp2}(\subref{exp2_1}), Fig. \ref{exp2}(\subref{exp2_3})) and prediction accuracy (Fig. \ref{exp2}(\subref{exp2_2}), Fig. \ref{exp2}(\subref{exp2_4})). The curve of CSR-$\{$VAE,CBM$\}$ in the prediction accuracy graph shows the prediction accuracy using all concepts, while the accuracy is shown in the other curve just using the selected concepts.

The result in Fig. \ref{exp2}(\subref{exp2_1}) and Fig. \ref{exp2}(\subref{exp2_3}) shows that the concept selection rate decreases by increasing $\alpha _5$ in our method, which implies that concept selection with concept sparsity regularization can keep the number of important concepts relatively small. When compared with VAE/Knockoff, it selects a large number of concepts 
even after using Knockoff. This means that concept sparsity regularization plays an important role in suppressing the number of the selected concept. Compared with $\beta-$VAE (red-dotted line), the number of selected concepts can be suppressed as CSR does use large $\beta$. 

The result in Fig. \ref{exp2}(\subref{exp2_2}) and Fig. \ref{exp2}(\subref{exp2_4})  shows that the prediction accuracy also increases with the increase of $\alpha _5$. This is because we set $\alpha _4 = 100 \times \alpha _5$. This setting means prediction error is more heavily penalized as $\alpha _5$ increases. In the unsupervised learning setting, the prediction accuracy of the models using all concepts and using concepts selected by Knockoff surpasses the accuracy of VAE and $\beta-$ VAE. At the same time, only the model using concepts selected by our method can achieve the same prediction accuracy as when all concepts are used, thus demonstrating that our method can make the concepts 
learned by $\hat{\phi}$ sparse and select out the concepts that are important for prediction. In the supervised learning setting, as $\alpha _5$ increases, the prediction accuracy when using only the concepts selected by our method and when using all concepts gradually shows a gap. The reason is that each concept has to be trained to correspond to a specific meaning, and an overly sparsified model will lead to a decrease in power, affecting the prediction accuracy.

\begin{figure}
    \centering
    \begin{subfigure}{.22\textwidth}
   \centering
    \includegraphics[width=\linewidth]{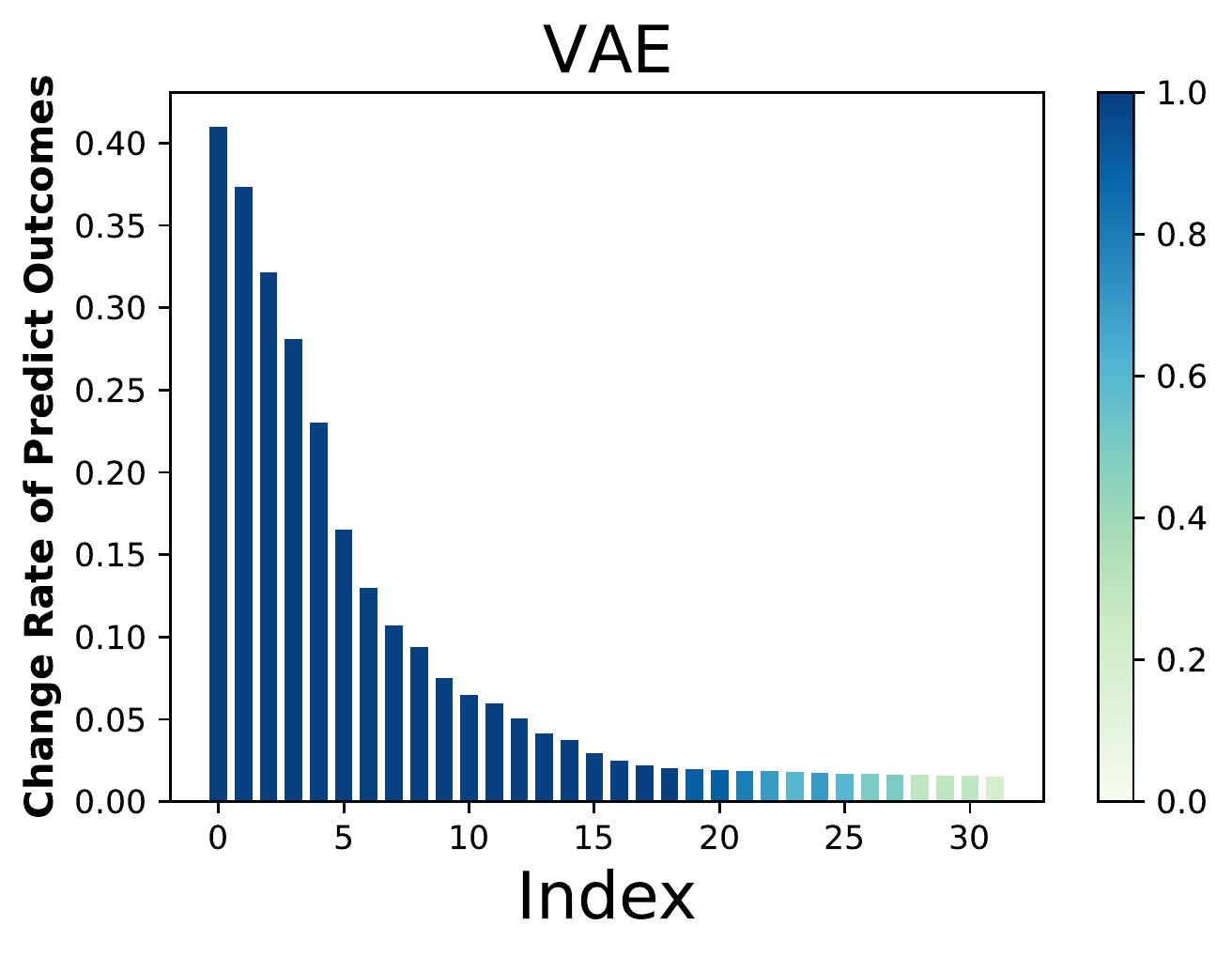}
    \subcaption{change rate (VAE)}
    \label{demo_VAE}
  \end{subfigure}
  \hfill
  \begin{subfigure}{.22\textwidth}
    \centering
    \includegraphics[width=\linewidth]{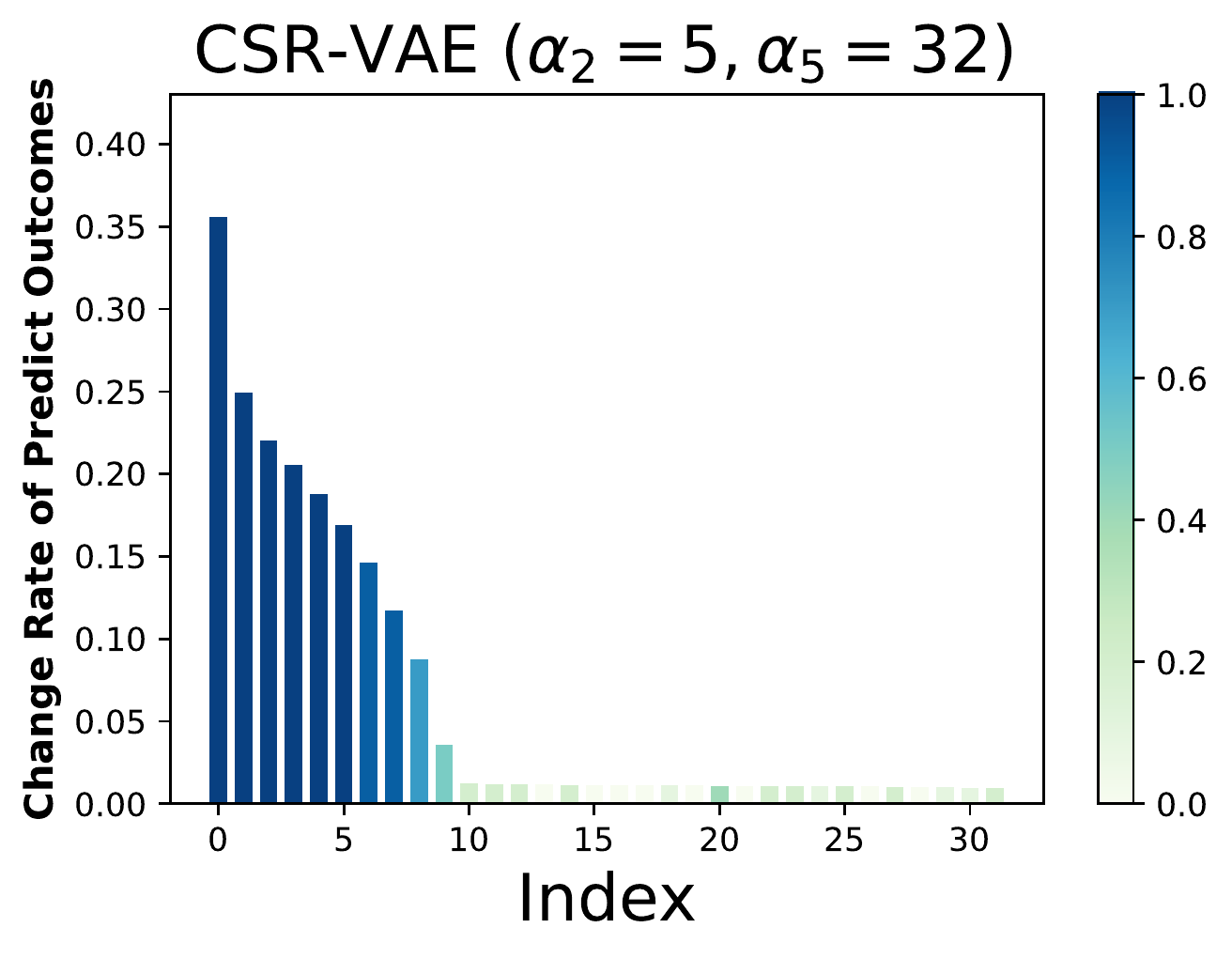}
    \subcaption{change rate (CSR-VAE)}
    \label{demo_CSR}
  \end{subfigure}%
  \caption{The prediction outcomes' change rate of reconstruction image by $\omega$ and feature selection results by different methods.}
  \label{exp_demo}
\end{figure}

\begin{figure}[t]
\begin{center}
\includegraphics[width=.5\textwidth]{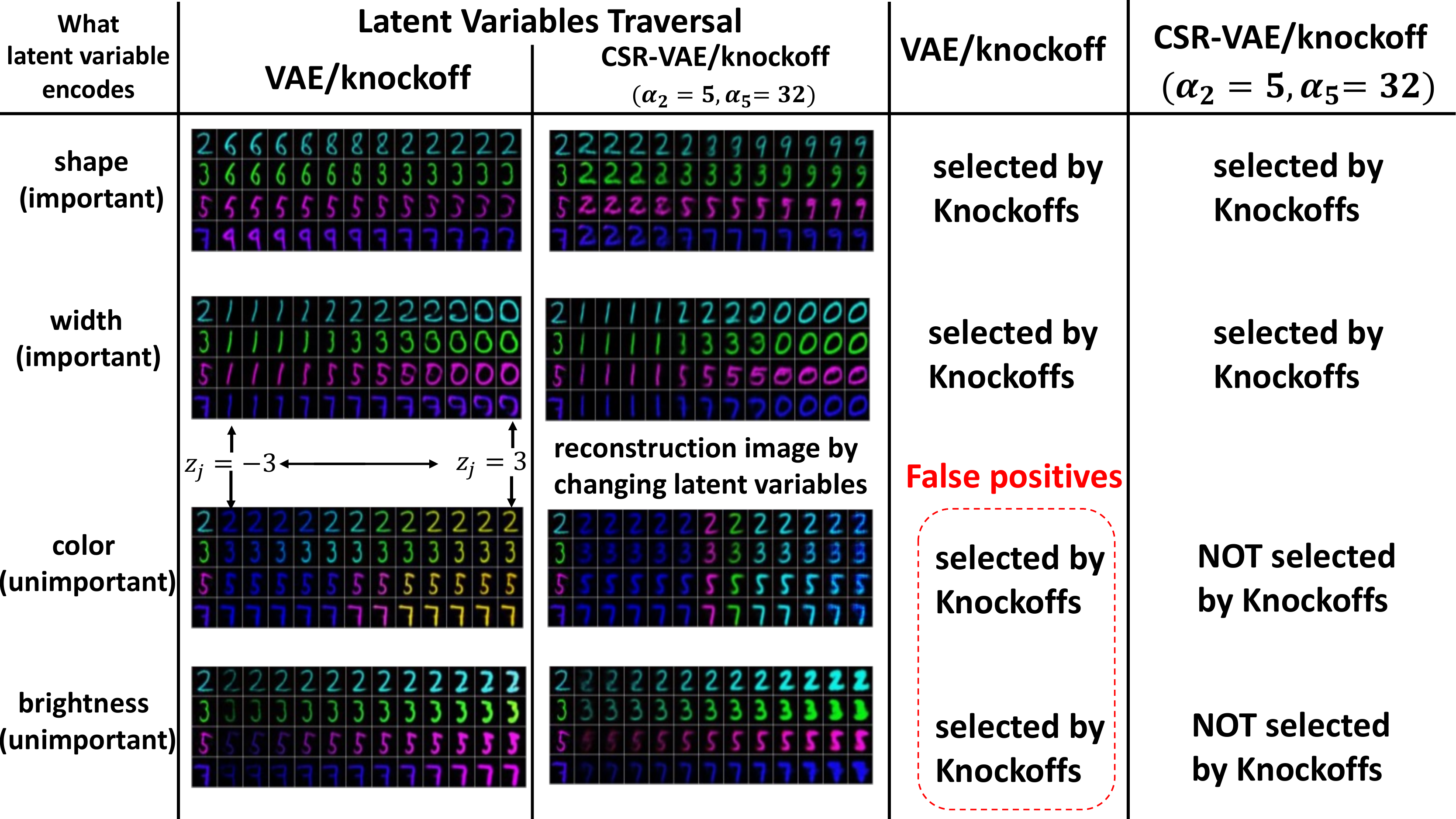}
\end{center}
\caption{Demonstration of feature selection for digit classification in the unsupervised setting (Colored-MNIST).}
\label{demo_expunsup}
\end{figure}

\begin{figure}[t]
\begin{center}
\includegraphics[width=.5\textwidth]{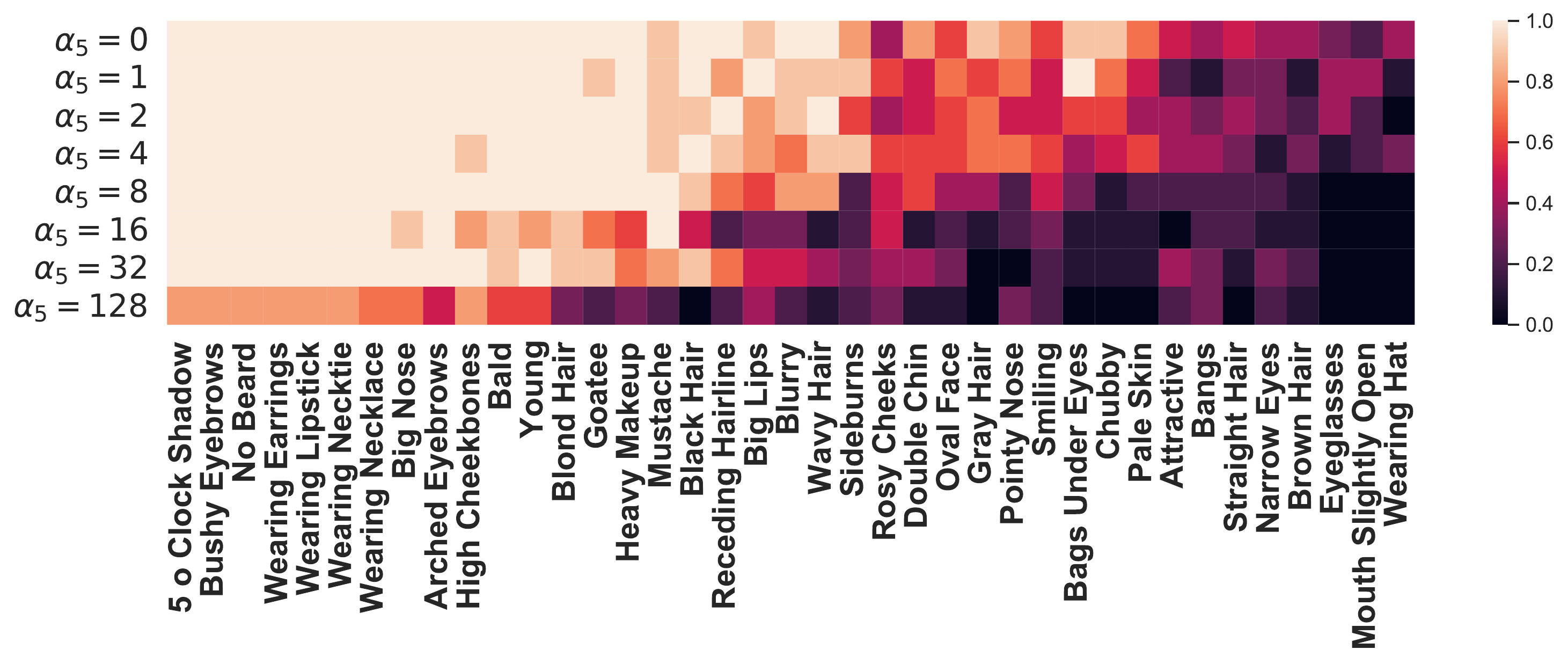}
\end{center}
\caption{Demonstration of feature selection results in the supervised setting (CelebA).}
\label{demo_expsup}
\end{figure}


\subsubsection{Results of Selected Concepts in the Unsupervised Setting}
In the unsupervised setting, we train a classifier $\omega$ on Colored MNIST to classify digits which helps us determine the important concepts. We intervene on each latent variable generated by VAE/CSR-VAE between [-3, 3] and use $\omega$ to predict the class of reconstruction images. We calculate the ratio by how much the reconstruction and original images class becomes different and denote it as the change rate. If a latent variable is considered important for prediction, it should have a relatively larger change rate value than other latents. 

Fig. \ref{exp_demo} shows the change rates and feature selection results by VAE/Knockoff and CSR-VAE/Knockoff. The horizontal axes represent the index of the latent variable, and the vertical axes represent the mean value of the prediction outcome's change rate by $\omega$ under $10$ independent runs and sorted by value. The value of the color bar represents the frequency that each latents selected by Knockoff among $10$ runs. We observe that the change rates of CSR-VAE are lower than VAE overall. This is because the distribution of $\bm{\mathrm{z}}$ generated by $\phi$ learned by different algorithms is different, which leads to the fact that for the same intervention, it may be able to affect the reconstruction of VAE but not CSR-VAE. Both Fig. \ref{exp_demo}(a) and Fig. \ref{exp_demo}(b) show that Knockoff is effective in selecting the latent variables to attain greater change rates. CSR-VAE/Knockoff can select a small number of important features for prediction, while VAE/Knockoff selects more than $80\%$ of features containing unimportant features. For demonstration purposes, we choose two important concepts (shape and width) and two unimportant concepts (color and brightness) among all concepts by their looks. In Fig. \ref{demo_expunsup}, the first column of the left side represents the original image, and the remaining columns show the decoder's reconstructions after intervention. The results show that VAE/Knockoff incorrectly treats color and brightness as important features. Even with latent variables with approximately the same meaning, the feature selection results by Knockoff can be different.  At the same time, our proposed CSR-VAE/Knockoff can effectively suppress these false positive explanations, improving the explanation's reliability.
\subsubsection{Results of Selected Concepts in the Supervised Setting} 
In Fig. \ref{demo_expsup}, each row represents the selection rate of each concept under different $10$ independent runs in CelebA. The concept is sorted by the mean value of the concept selection rate under all $\alpha _5$ (\textit{i.e.,} the left side concepts is more likely to be selected under all hyperparameters). The method that gives the result in the top row is equal to CBM/Knockoff and each of the remaining rows corresponds to CSR-CBM/Knockoff with varying $\alpha _5$. We observe that the selection rate of each concept decreases by increasing the value of $\alpha _5$. When $\alpha _5$ increases, concepts show a decreasing trend in selection rate, which effectively suppresses the occurrence of false positives. We found that concepts on the left side are subjectively important in human vision, such as “no beard” or “wearing earrings”.
While some subjectively unimportant concepts are also selected with a high probability (\textit{e.g.,} “young”, “blond hair”). The possible reason is that there is some bias in the important features used by the model and the human in making predictions, which leads to some subjective false positive concepts that may be important for the model to predict.

\section{Conclusion}
In summary, we proposed a method to suppress the false positive concept-based explanation by controlling the FDR of selected concepts under a certain level. Also, we propose a CSR loss that can be added to the concept model to more effectively apply the Knockoff filter for concept selection and thus improve the interpretability and reliability of the model. 


\section*{Acknowledgments}
This work is partly supported by Japan science and technology agency (JST), CREST JPMJCR21D3, and Japan society for the promotion of science (JSPS), Grants-in-Aid for Scientific Research 23H00483,  22H00519, and 22H00521.

\bibliographystyle{named}
\bibliography{ijcai23}

\end{document}